\pdfoutput=1

\documentclass[11pt]{article}

\usepackage[]{ACL2023}

\usepackage{times}
\usepackage{latexsym}

\usepackage[T1]{fontenc}

\usepackage[utf8]{inputenc}

\usepackage{microtype}

\usepackage{inconsolata}
\usepackage{graphicx}
\usepackage{amsmath}
\usepackage{color}
\usepackage{multirow}
\usepackage{setspace}
\usepackage{tabularx}
\usepackage{wrapfig, lipsum, booktabs}
\usepackage{arydshln}
\usepackage{bbding}
\usepackage{dingbat}
\usepackage{pifont} 
\usepackage{wasysym}
\usepackage{amssymb}
\newcommand{\okmark}{{\textbf{\color[rgb]{0.1, 0.5, 0.1}\ding{51}}}}
\newcommand{\ngmark}{{\textbf{\color{red}{\ding{55}}}}}

\title{\textsc{Decker}: Double Check with Heterogeneous Knowledge \\
for Commonsense Fact Verification}

\author{Anni Zou\textsuperscript{1,2}, Zhuosheng Zhang\textsuperscript{1,2}, Hai Zhao\textsuperscript{1,2,\thanks{\ \ Corresponding author. This paper was partially supported by Key Projects of National Natural Science Foundation of China (U1836222 and 61733011).}} \\
\textsuperscript{1} Department of Computer Science and Engineering, Shanghai Jiao Tong University\\
\textsuperscript{2} Key Laboratory of Shanghai Education Commission for Intelligent Interaction\\
and Cognitive Engineering, Shanghai Jiao Tong University\\
\texttt{\{annie0103,zhangzs\}@sjtu.edu.cn,zhaohai@cs.sjtu.edu.cn}\\
}

\begin{document}
\maketitle

\begin{abstract}
Commonsense fact verification, as a challenging branch of commonsense question-answering (QA), aims to verify through facts whether a given commonsense claim is correct or not. Answering commonsense questions necessitates a combination of knowledge from various levels. However, existing studies primarily rest on grasping either unstructured evidence or potential reasoning paths from structured knowledge bases, yet failing to exploit the benefits of heterogeneous knowledge simultaneously. In light of this, we propose \textsc{Decker}, a commonsense fact verification model that is capable of bridging heterogeneous knowledge by uncovering latent relationships between structured and unstructured knowledge. Experimental results on two commonsense fact verification benchmark datasets, CSQA2.0 and CREAK demonstrate the effectiveness of our \textsc{Decker} and further analysis verifies its capability to seize more precious information through reasoning. The official implementation of \textsc{Decker} is available at \url{https://github.com/Anni-Zou/Decker}.
\end{abstract}

\section{Introduction}
\begin{figure}
    \centering
    \includegraphics[width=0.5\textwidth]{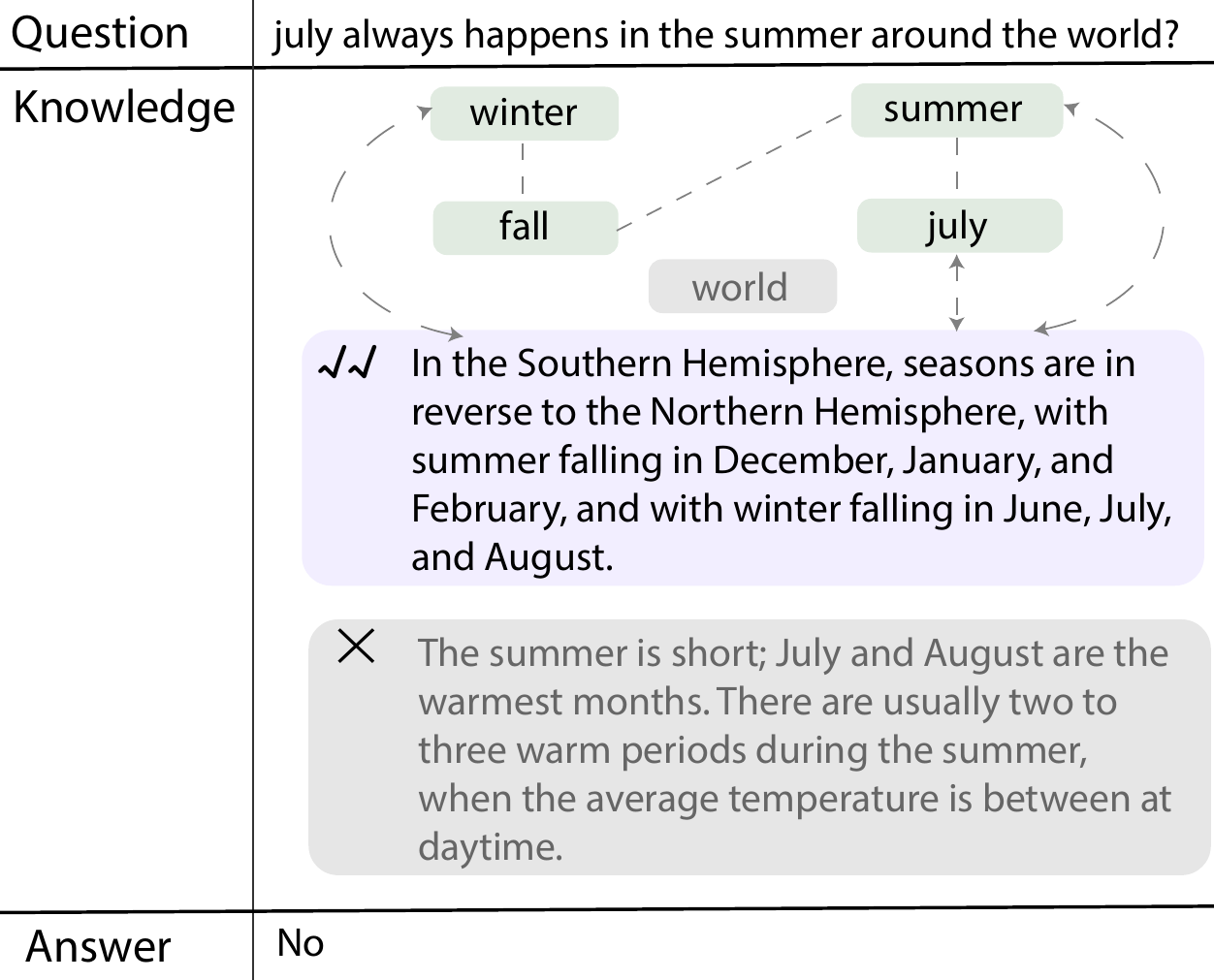}
    \caption{An example from CSQA2.0 \citep{talmor2022commonsenseqa}. Given the question, we perform a double check between the heterogeneous knowledge (i.e., KG and facts) and aim to derive the answer by seizing the valued information through reasoning.}
    \label{example}
\end{figure}

Commonsense question answering is an essential task in question answering (QA), which requires models to answer questions that entail rich world knowledge and everyday information. The major challenge of commonsense QA is that it not only requires rich background knowledge about how the world works,  but also demands the ability to conduct effective reasoning over knowledge of various types and levels \citep{hudson2018compositional}. Recently, there emerges a challenging branch of commonsense QA: commonsense fact verification, which aims to verify through facts whether a given commonsense claim is correct or not \citep{onoe2021creak, talmor2022commonsenseqa}. Different from previous \textit{multiple-choice} settings which contain candidate answers \citep{talmor-etal-2019-commonsenseqa}, commonsense fact verification solely derives from the question itself and implements reasoning on top of it (Figure \ref{example}). Therefore, it poses a novel issue of how to effectively seize the useful and valuable \textit{knowledge} to deal with commonsense fact verification.

One of the typical methods is to make direct use of knowledge implicitly encoded in pre-trained language models (PLMs) \citep{devlin-etal-2019-bert, liu2019roberta, he2021debertav3}, which have proved to be useable knowledge bases \citep{petroni2019language, bosselut-etal-2019-comet}. The knowledge in PLMs is gained during the pre-training stage through mining large-scale collection of unstructured text corpora. Nevertheless, the sore spot lies in that it is natural for human brains to project our prior world knowledge onto the answers facing the commonsense questions \citep{lin-etal-2019-kagnet, choi}, whereas it is tough for PLMs to learn commonsense knowledge that is implicitly stated in plain texts from corpora \citep{gunning2018machine}.

To strengthen PLMs to perform commonsense QA, there is a surging trend of methods equipping language models with different levels of external knowledge, encompassing structured knowledge such as knowledge graphs (KG) \citep{lin-etal-2019-kagnet, yan-etal-2021-learning, yasunaga-etal-2021-qa, zhang2022greaselm} and unstructured knowledge such as text corpus \citep{lin-etal-2021-differentiable, yu2022retrieval}. While the KG-based methods yield remarkable performances on commonsense QA recently, they are more suitable and adaptive for \textit{multiple-choice} settings because they lay emphasis on discovering connected patterns between the question and candidate answers. For example, to answer a question \textit{crabs live in what sort of environment?} with candidate answers \textit{saltwater}, \textit{galapagos} and \textit{fish market}, the KG-based methods manage to capture the path \textit{crab--sea--saltwater} in KG, leading to a correct prediction. Nonetheless, they encounter a bottleneck when dealing with commonsense fact verification. Figure \ref{example} shows an example: when asked whether \textit{july always happens in the summer around the worlds}, the KG-based methods have a tendency to detect a strong link between \textit{july} and \textit{summer}, which may persuade the model to deliver the wrong prediction. 

In general, there are two major limitations in previous studies. On one hand, structured knowledge abounds with structural information among the entities but suffers from sparsity and limited coverage. On the other hand, unstructured knowledge provides rich and broad context-aware information but undergoes noisy issues. These two kinds of knowledge can be naturally complementary to each other. However, most existing works focus on either structured or unstructured external knowledge but fail to exploit the benefits of heterogenous knowledge simultaneously. As the example in Figure \ref{example} shows: if we rely only on the structured knowledge in KG, we tend to derive that \textit{july} and \textit{summer} are strongly correlated, with an extremely weak relationship between \textit{summer} and \textit{winter}. Similarly, if we focus only on the textual facts, we are more inclined to focus on the fact in grey, as it describes more information about \textit{summer} in \textit{july}. As a consequence, uncovering latent relationships among heterogeneous knowledge helps bridge the gap and yield more valuable and useful information.

Motivated by the above ideas, we propose \textsc{Decker}, a commonsense fact verifier that bridges heterogeneous knowledge and performs a double check based on interactions between structured and unstructured knowledge. Our proposed \textsc{Decker} works in the following steps: (i) firstly, it retrieves heterogeneous knowledge including a KG subgraph and several relevant facts following prior works \citep{zhang2022greaselm, izacard2022unsupervised}; (ii) secondly, it constructs an integral graph with encoded question and facts and then employs relational graph convolutional networks (R-GCN) to reason and filter over the heterogenous knowledge; (iii) lastly, it adopts a multi-head attention pooling mechanism to obtain a final refinement of enriched knowledge representation and combines it with the question representation for downstream tasks.

Our contributions are summarized as follows:

(i) For the concerned commonsense fact verification task, we initialize the research that simultaneously takes heterogeneous knowledge into account.

(ii) We propose a novel method in terms of R-GCN to construct an integral graph that executes a double check between structured and unstructured knowledge and better uncovers the latent relationships between them.

(iii) Experimental results on two commonsense fact verification benchmarks show the effectiveness of our approach, verifying the necessity and benefits of heterogeneous knowledge integration.

\section{Related Work}
\subsection{Commonsense QA}
Commonsense QA is a long-standing challenge in natural language processing as it calls for intuitive reasoning about real-world events and situations \citep{csr}. As a result, recent years have witnessed a plethora of research on developing commonsense QA tasks, including SWAG \citep{zellers-etal-2018-swag}, Cosmo QA \citep{huang-etal-2019-cosmos}, HellaSwag \citep{zellers-etal-2019-hellaswag}, CSQA \citep{talmor-etal-2019-commonsenseqa}, SocialIQa \citep{sap-etal-2019-social} and PIQA \citep{bisk2020piqa}. However, these tasks primarily attend to \textit{multiple-choice} settings, so that there usually exist potential reasoning paths  which explicitly connect the question with candidate answers. This may cause the models to be susceptible to shortcuts during reasoning \citep{zhang2022greaselm}. Therefore, a novel branch of commonsense QA: commonsense fact verification has emerged to further exploit the limits of reasoning models, such as CREAK \citep{onoe2021creak} and CSQA2.0 \citep{talmor2022commonsenseqa}. Unlike previous \textit{multiple-choice} settings, commonsense fact verification needs the models to be granted richer background knowledge and higher reasoning abilities based on the question alone. Hence, our work dives into commonsense fact verification and conducts experiments on two typical benchmarks: CREAK and CSQA2.0.

\subsection{Knowledge-enhanced Methods for Commonsense QA}
Despite the impressive performance of PLMs on many commonsense QA tasks, they struggle to capture sufficient external world knowledge about concepts, relations and commonsense \citep{zhu-etal-2022-knowledge}. Therefore, it is of crucial importance to introduce external knowledge for commonsense QA. Currently, there are two major lines of research based on the property of knowledge: structured knowledge (i.e., knowledge graphs) and unstructured knowledge (i.e., text corpus).

The first research line strives to capitalize on distinct forms of knowledge graphs (KG), such as Freebase \citep{freebase}, Wikidata \citep{wikidata}, ConceptNet \citep{speer2017conceptnet}, ASCENT \citep{nguyen2021advanced} and ASER \citep{zhang2022aser}. Commonsense knowledge is thus explicitly delivered in a triplet form with relationships between entities. An initial thread of works endeavors to discover potential reasoning paths between the question and candidate answers under \textit{multiple-choice} settings, which have shown remarkable advances in structured reasoning and question answering. For example, KagNet \citep{lin-etal-2019-kagnet} utilizes a hierarchical path-based attention mechanism and graph convolutional networks to cope with relational reasoning. MHGRN \citep{feng-etal-2020-scalable} modifies from graph neural networks to make it adaptable for multi-hop reasoning while HGN \citep{yan-etal-2021-learning} conducts edge generation and reweighting to find suitable paths more efficiently. JointLK \citep{sun-etal-2022-jointlk} performs joint reasoning between LM and GNN and uses the dynamic KGs pruning mechanism to seek effective reasoning. Furthermore, other research optimizes by enhancing the interaction between raw texts of questions and KG to achieve better performance and robustness. QA-GNN \citep{yasunaga-etal-2021-qa} designs a relevance scoring to make the interaction more effective, whereas GreaseLM \citep{zhang2022greaselm} leverages multiple layers of modality interaction operations to achieve deeper interaction. Nevertheless, the scope of commonsense knowledge is infinite, far beyond a knowledge graph defined by a particular pattern. 

The second research line attempts to make use of unstructured knowledge with either prompting methods \citep{lal-etal-2022-using, qiao2023reasoning} or information retrieval techniques \citep{retrieve}.
Maieutic prompting \citep{jung-etal-2022-maieutic} infers a tree of explanations through abductive and recursive prompting from generations of large language models (LLMs), which incurs high inference costs due to paywalls imposed by LLMs providers. DrFact \citep{lin-etal-2021-differentiable} retrieves the related facts step by step through an iterative process of differentiable operations and further enhances the model with an external ranker. 
\citet{Talmor2020TeachingPM} employs regenerated data to train the model to reliably perform systematic reasoning. RACo \citep{yu2022retrieval} utilizes a \textit{retriever-reader} architecture as the backbone and retrieves documents from a large-scale mixed commonsense corpus. \citet{xu-etal-2021-fusing} extracts descriptions of related concepts as additional input to PLMs. However, these works mainly focus on homogeneous knowledge and reason on top of it, ignoring the need to fuse multiple forms of knowledge. Unlike previous works, our model is dedicated to intuitively modeling the relations between heterogeneous knowledge, bridging the gap between them, and filtering the more treasured knowledge by exploiting their complementary nature, in an inference-cost-free pattern. 

Besides, there are some works taking heterogeneous knowledge into account to deal with commonsense reasoning. For instance, \citet{lin-etal-2017-reasoning} mines various types of knowledge (including event narrative knowledge, entity semantic knowledge and sentiment coherent knowledge) and encodes them as inference rules with costs to tackle commonsense machine comprehension. Nevertheless, this work is principally based on semantic or sentiment analysis at the sentence level, seeking knowledge enrichment at various levels of granularity. Our approach, however, is more concerned with extending external sources of knowledge and creating connections between heterogeneous knowledge from distinct sources so that they may mutually filter each other.

\begin{figure*}[htb]
    \centering
    \includegraphics[width=1.0\textwidth]{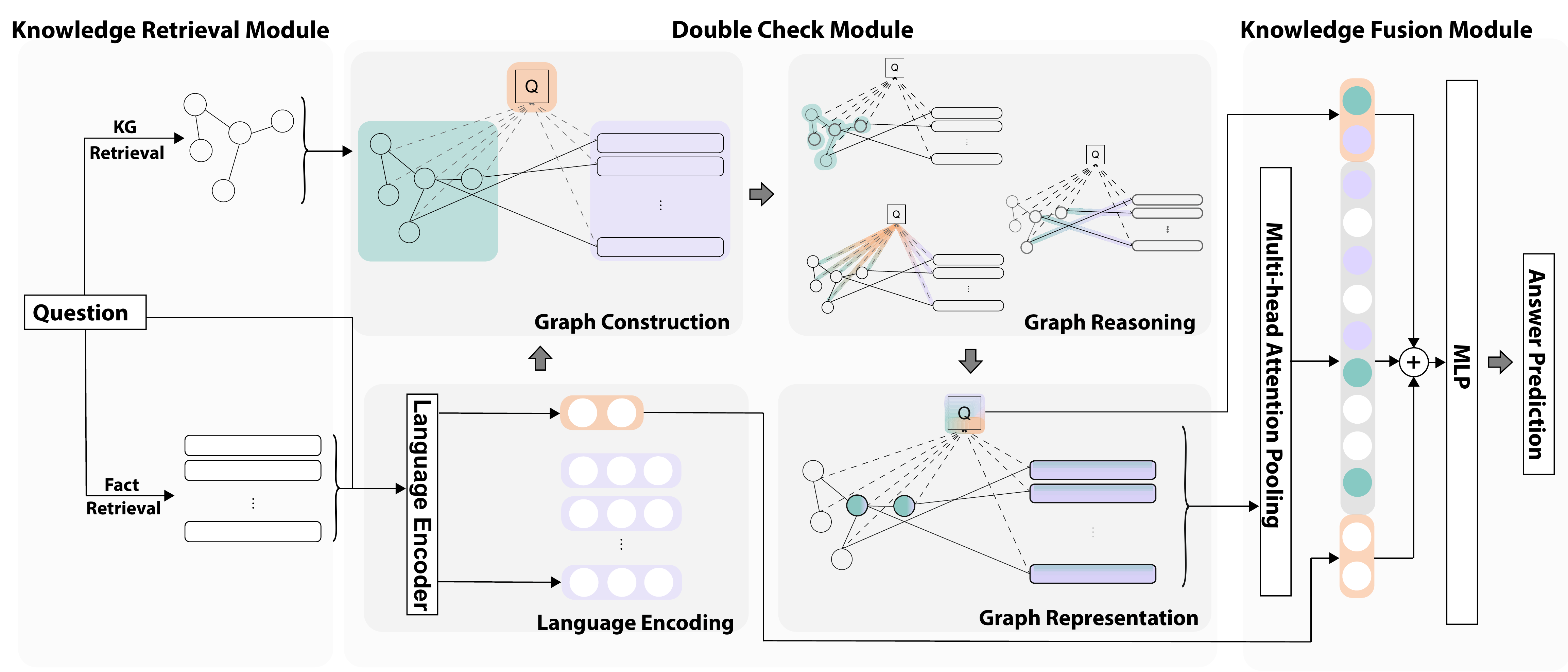}
    \caption{Overview of our approach, which consists of three components: Knowledge Retrieval Module (left), Double Check Module (middle), and Knowledge Fusion Module (right). Given an input question, KG retriever and fact retriever extract relevant local KG and facts (Knowledge Retrieval Module); then heterogeneous knowledge including entities in KG and facts are enhanced (Double Check Module); finally, heterogeneous knowledge is merged to deduce the final answer prediction (Knowledge Fusion Module).}
    \label{overview}
\end{figure*}

\section{Methodology}
This section presents the details of our proposed approach. Figure \ref{overview} gives an overview of its architecture. Our approach, \textsc{Decker}, consists of three major modules: (i) Knowledge Retrieval Module which retrieves heterogeneous knowledge based on the input question; (ii) Double Check Module which merges information from structured and unstructured knowledge and makes a double check between them; (iii) Knowledge Fusion Module which combines heterogeneous knowledge together to obtain a final representation.

\subsection{Knowledge Retrieval Module}
\paragraph{KG Retriever}
Given a knowledge graph $\mathcal{G}$ and an input question $q$, the goal of the KG Retriever is to retrieve a question-related sub-graph $\mathcal{G}_{sub}^{q}$ from $\mathcal{G}$. Following previous works \citep{lin-etal-2019-kagnet, yasunaga-etal-2021-qa, zhang2022greaselm}, we first execute entity linking to $\mathcal{G}$ to extract an initial set of nodes $\mathcal{V}_{init}$. We then obtain the set of retrieved entities $\mathcal{V}_{sub}$ by adding any bridge entities that are in a 2-hop path between any two linked entities in $\mathcal{V}_{init}$. Eventually, the retrieved subgraph $\mathcal{G}_{sub}$ is formed by retrieving all the edges that join any two nodes in $\mathcal{V}_{sub}$.

\paragraph{Fact Retriever}
Given a large corpus of texts containing $K$ facts and an input question $q$, the objective of the fact retriever is to retrieve the top-$k$ facts relevant to $q$. Following Contriever \citep{izacard2022unsupervised} which is an information retrieval model pre-trained using the MoCo contrastive loss \citep{he2020momentum} and unsupervised data only, we employ a dual-encoder architecture where the question and facts are encoded independently by a BERT base uncased model \citep{huang2013learning,karpukhin-etal-2020-dense}. For each question and fact, we apply average pooling over the outputs of the last layer to obtain its corresponding representation. Then a relevance score between a question and a fact is obtained by computing the dot product between their corresponding representations. 

More precisely, given a question $q$ and a fact $f_i \in\left\{f_1, f_2, \ldots, f_K\right\}$, we encode each of them independently using the same model. The relevance score $r(q,f_{i})$ between a question $q$ and a fact $f_{i}$ is the dot product of their resulting representations:
\begin{equation}
    r(q, f_{i})=\left\langle E_\theta(q), E_\theta(f_{i})\right\rangle,
\end{equation}
where $\langle, \rangle$ denotes the dot product operation and $E_{\theta}$ denotes the model parameterized by $\theta$.

After obtaining the corresponding relevance scores, we select $k$ facts $\mathcal{F}=\left\{f_q^{1}, f_q^{2},\ldots,f_q^{k}\right\}$, whose relevance scores $r(q,f)$ are top-$k$ highest among all $K$ facts for each question $q$.

\subsection{Double Check Module}\label{sec:double}
\paragraph{Language Encoding}
Given a question $q$ and a set of retrieved facts $\mathcal{F}=\left\{f_q^{1}, f_q^{2},\ldots,f_q^{k}\right\}$, we deliver their corresponding sets of tokens $\mathcal{Q}=\left\{q^1, q^2, \ldots, q^t\right\}$ and $f_q^i = \left\{t_i^1, t_i^2, \ldots, t_i^{o_i}\right\}$ into a PLM, where $t$ and $o_{i}$ are the lengths of the question and fact sequence $f_q^i$, respectively. We obtain their representations independently by extracting \texttt{[CLS]} inserted at the beginning:
\begin{equation}
    \begin{split}
        q_{enc} &= \text{Encoder}\left(\left\{q^1, q^2, \ldots, q^t\right\}\right) \in \mathcal{R}^{d},\\
        f_{enc}^i &= \text{Encoder}\left(\left\{t_i^1, t_i^2, \ldots, t_i^{o_i}\right\}\right) \in \mathcal{R}^{d}, \\
        \mathcal{F}_{enc} &= \left\{f_{enc}^1, f_{enc}^2, \ldots, f_{enc}^{k}\right\} \in \mathcal{R}^{k \times d},
    \end{split}
    \label{encoder}
\end{equation}
where $d$ denotes the hidden size defined by PLM.

\begin{figure}[t]
\centering
\includegraphics[width=0.48\textwidth]{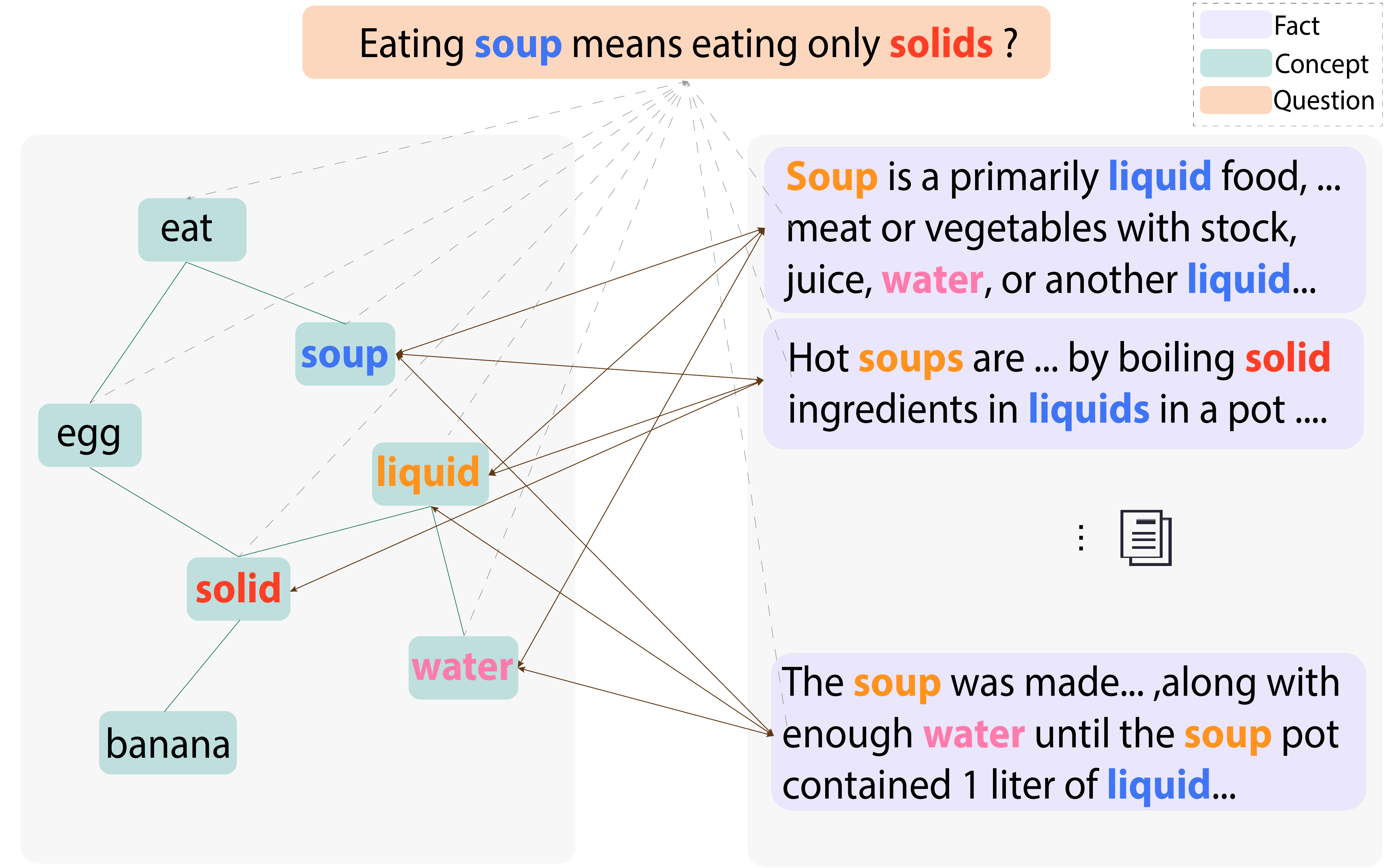}
\caption{An example of the constructed integral graph.}
\label{graph}
\end{figure}

\paragraph{Graph Construction}
Figure \ref{graph} gives an example of the constructed graph, which is dubbed as \textit{integral graph}. Given a question $q$, a sub-graph $\mathcal{G}_{sub}^{q}$ extracted from KG and several retrieved facts $\mathcal{F}=\left\{f_q^{1}, f_q^{2},\ldots,f_q^{k}\right\}$, we construct an integral graph denoted as $\mathcal{G}=(\mathcal{V},\mathcal{E},\mathcal{R})$. Here $\mathcal{V} = \mathcal{V}_q \cup \mathcal{V}_c \cup \mathcal{V}_f$ is the set of entity nodes, where $\mathcal{V}_q$, $\mathcal{V}_c$ and $\mathcal{V}_f$ denote the \textit{question node} (\textbf{\textcolor[RGB]{252,175,124}{orange}} in Figure \ref{graph}), \textit{concept nodes} (\textbf{\textcolor[RGB]{135,201,195}{green}} in Figure \ref{graph}) and \textit{fact nodes} (\textbf{\textcolor[RGB]{222,213,255}{purple}} in Figure \ref{graph}), respectively; $\mathcal{E} $ is the set of edges that connect nodes in $\mathcal{V}$; $\mathcal{R}$ is a set of relations representing the type of edges in $\mathcal{E}$. In the integral graph, we define four types of edges\footnote{We ignore fact-to-fact edges due to the reason that if a fact-to-fact edge is added when the two facts link to the same concept node, a performance drop will be observed on the CREAK dev set (89.5\% -> 87.3\%).}: 
 
$\bullet$ \quad concept-to-fact edges: $(n_c, r_{c2f}, n_f)$; 

$\bullet$ \quad concept-to-concept edges: $(n_c, r_{c2c}, n_c)$;

$\bullet$ \quad question-to-fact edges: $(n_q, r_{q2f}, n_f)$;

$\bullet$ \quad question-to-concept edges: $(n_q, r_{q2c}, n_c)$,
where $n_q \in \mathcal{V}_q$, $n_c \in \mathcal{V}_c$, $n_f \in \mathcal{V}_f$ and $\left\{r_{c2f}, r_{c2c}, r_{q2f}, r_{q2c}\right\} \subseteq \mathcal{R}$.

For question-to-concept and question-to-fact edges which are bidirectional, we connect the question node with all the other nodes in the integral graph with regard to enhancing the information flow between the question and its related heterogeneous knowledge. For concept-to-concept edges which are directional, we keep the structured knowledge extracted from KG and do not distinguish the multiple relations inside the sub-graph, as our approach mainly concentrates on effective reasoning over heterogeneous knowledge. For concept-to-fact edges, we use string matching and add a bidirectional edge $(n_{c}, r_{c2f}, n_{f})$ between $n_c \in \mathcal{V}_c$ and $n_f \in \mathcal{V}_f$ with $r_{c2f} \in \mathcal{R}$ if the concept $n_c$ can be captured in the fact $n_f$. For instance, there should exist an edge between the concept \textit{\underline{soup}} and the fact \textit{\underline{soup} is primarily a liquid food}. In this way, the noisy and peripheral information is filtered whereas the relevant and precious knowledge is intensified.

Afterward, we initialize the node embeddings in the integral graph $\mathcal{G}$. For the concept nodes, we follow the method of prior work \citep{feng-etal-2020-scalable, zhang2022greaselm} and employ pre-trained KG embeddings for the matching nodes, which is introduced in Section \ref{kg}. Then the pre-trained embeddings go through a linear transformation to align the dimension:
\begin{equation}
    \begin{split}
        & \mathcal{C}_{emb} = \left\{c^1, c^2, 
 \ldots, c^m\right\} \in \mathcal{R}^{m \times d_c}, \\
        & \mathcal{C}_{graph} = \mathcal{C}_{emb} W_{c} + b_c \in \mathcal{R}^{m \times d},
    \end{split}
    \label{cpt_emb}
\end{equation}
where $m$ denotes the number of concept nodes in the sub-graph, $d_c$ denotes the hidden size of pre-trained KG embeddings, $W_{c} \in \mathcal{R}^{d_c \times d}$ and $b_c \in \mathcal{R}^d$ are trainable transformation matrices and bias vectors respectively.

For the question nodes and fact nodes, we inject the corresponding encoded results from PLM in Equation \ref{encoder}. Consequently, we obtain the initial node embeddings $\mathcal{N}^{(0)} \in \mathcal{R}^{(1+k+m) \times d}$ for the integral graph:
\begin{equation}
    \mathcal{N}^{(0)} = \left[{q_{enc}}^{(0)}; {\mathcal{F}_{enc}}^{(0)};{\mathcal{C}_{graph}}^{(0)}\right].
\end{equation}

\paragraph{Graph Reasoning}
As our integral graph $\mathcal{G}$ is a multi-relational graph where distinct edge types serve as varied information exchange between disparate knowledge, the message-passing process from a source node to a target node should be aware of its relationship, \textit{i.e.,} relation type of the edge. For example, the concept-to-fact edges help to implement a double check and filtering between concepts and facts whereas the concept-to-concept edges assist in discovering the structured information. To this end, we adopt relational graph convolutional network (R-GCN) \citep{schlichtkrull2018modeling} to perform reasoning on the integral graph. 

In each layer of R-GCN, the current node representations $\mathcal{N}^{(l)}$ are fed into the layer to perform a round of information propagation between nodes in the graph and yield novel representations:
\begin{equation}
    \mathcal{N}^{(l+1)} = \text{R-GCN} \left(\mathcal{N}^{(l)}\right).
\end{equation}

More precisely, the R-GCN computes node representations $h_i^{(l+1)} \in \mathcal{N}^{(l+1)}$ for each node $n_i \in \mathcal{V}$ by accumulating and inducing features from neighbors via message passing:
\begin{equation}
    h_i^{(l+1)}=\sigma\left(\sum_{r \in \mathcal{R}} \sum_{j \in N_i^r} \frac{1}{c_{i, r}} W_r^{(l)} h_j^{(l)}+W_0^{(l)} h_i^{(l)}\right),
\end{equation}
where $\mathcal{R}$ is the set of relations, which corresponds to four edge types in our integral graph. $N_i^r$ denotes the set of neighbors of node $n_i$, which are connected to $n_i$ under relation $r$, and $c_{i,r}$ is a normalization constant. $W_r^{(l)}$ and $W_0^{(l)}$ are trainable parameter matrices of layer $l$. $\sigma$ is an activated function, which in our implementation is GELU \citep{hendrycks2016gaussian}.

Finally, we access the graph output through an $L$-layer R-GCN:
\begin{equation}
    N^{(L)} = \left[{q_{enc}}^{(L)}; {\mathcal{F}_{enc}}^{(L)}; {\mathcal{C}_{graph}}^{(L)}\right].
\end{equation}

\subsection{Knowledge Fusion Module}
\paragraph{Multi-head Attention Pooling}
Since the acquired heterogeneous knowledge is leveraged to help answer the question, further interaction between the question and the knowledge is needed to refine the double-checked knowledge. Following the idea of \citet{zhang2022greaselm}, we introduce a multi-head attention pooling mechanism (MHA) to ulteriorly gather the question-related information:
\begin{equation}
    \begin{split}
        & \text{Attn}(Q,K,V) = \operatorname{softmax}\left(\frac{Q K^T}{\sqrt{{d}_k}}\right) V,  \\
        & \text{head}_t = \text{Attn}\left(H_q W_t^Q, H_k W_t^K, H_k W_t^V\right), \\
        & \text{MHA}(H_q, H_k) = \left[\text{head}_1, \ldots, \text{head}_N\right] W^O, \\
    \end{split}
    \label{MHA}
\end{equation}
where $W_t^Q \in \mathcal{R}^{d \times d_q}$, $W_t^K \in \mathcal{R}^{d \times d_k}$, $W_t^V \in \mathcal{R}^{d \times d_v}$, $W^O \in \mathcal{R}^{hd_v \times d}$ are trainable parameter matrices, $h$ is the number of attention heads. $d_q$, $d_k$, $d_v$ denote the hidden sizes of the query vector, key vector and value vector, respectively.

Specifically, we employ the initial question embedding from PLM as the query and feed it into MHA together with the graph-encoded representations of facts and concepts \footnote{We use the initial question embedding from PLM because it can capture the original information about the question. To verify this, the query in MHA is replaced with the post-RGCN representation and a slight performance drop is observed (89.5\% -> 89.2\%) on the CREAK dev set.}. We thus derive the pooled knowledge representation:
\begin{equation}
    K_a = \text{MHA}\left(q_{enc}, \ \left[\mathcal{F}_{enc}^{(L)}; \mathcal{C}_{graph}^{(L)}\right]\right) \in \mathcal{R}^{d}.
\end{equation}

\paragraph{Answer Prediction}
In the end, we concatenate the initial question embeddings $q_{enc}$, the pooled knowledge representation $K_a$ and the enriched question representation $q_{enc}^{(L)}$ and deliver it into a predictor to get a final answer prediction:
\begin{equation}
    l = \text{MLP}\left([q_{enc}; K_a; q_{enc}^{(L)}]\right) \in \mathcal{R},
\end{equation}
where the predictor is a two-layer MLP with a tanh activation of size $(3d, d, nlabel)$,  $nlabel$ denotes the number of labels, which equals to $2$ in our commonsense fact verification setting. The model is optimized using the cross entropy loss.

\begin{table*}
    \centering
    \setlength{\tabcolsep}{15pt}
    \small
    \begin{tabular}{lccccc}
        \toprule
        \multirow{2}{*}{Model} & \#Total & Single-task  &\multicolumn{2}{c}{CREAK} & \multicolumn{1}{l}{CSQA2.0}\\
        & Params. & Training & Test & Contra & Test \\ 
        \midrule
        Human \citep{onoe2021creak} & & & - & 92.2 & - \\
        \cdashline{1-6}[0.6pt/2pt]
        GreaseLM \citep{zhang2022greaselm} & $\sim$359M & \okmark & 77.5 & - & - \\
        UNICORN \citep{lourie2021unicorn} & $\sim$770M  & \ngmark & 79.5 & - & 54.9 \\
        T5-3B \citep{t5} & $\sim$ 3B  &  \ngmark & 85.1 & 70.0 & 60.2 \\
        RACo \citep{yu2022retrieval} & $\geq$ 3B  & \ngmark & \textbf{88.6} & 74.4 & 61.8\\
        \midrule
        \textsc{Decker} (\textbf{Ours})  & $\sim$449M & \okmark & 88.4 & \textbf{79.2} & \textbf{68.1}\\
        \bottomrule
    \end{tabular}
    \caption{Experimental results on the CREAK and CSQA2.0 datasets. The evaluation metric is accuracy (acc).}\label{results}
\end{table*}

\section{Experiments}
\subsection{Datasets}
We conduct the experiments on two commonsense fact verification datasets: CommonsenseQA2.0 \citep{talmor2022commonsenseqa} and CREAK \citep{onoe2021creak}. The metric for evaluation is accuracy (acc).

\textbf{CommonsenseQA2.0} is a commonsense reasoning dataset collected through gamification. It includes 14,343 assertions about everyday commonsense knowledge. We use the original \textit{train / dev / test} splits from \citet{talmor2022commonsenseqa}.

\textbf{CREAK} is a dataset for commonsense reasoning about entity knowledge. It is made up of 13,000 English assertions encompassing 2,700 entities that are either true or false, in addition to a small contrast set. Each assertion is generated by a crowdworker based on a Wikipedia entity, which can be named entities, common nouns and abstract concepts. We perform our experiments using the \textit{train / dev / test / contrast} splits from \citet{onoe2021creak}.

\subsection{Experimental Setup}
\paragraph{Retrieval Corpus}
We leverage the English Wikipedia dump as the retrieval corpus. For preprocessing Wikipedia pages, we utilize the same method as described in \citet{karpukhin-etal-2020-dense, lewis2020retrieval}. We divide each Wikipedia page into separate 100-word paragraphs, amounting to 21,015,324 facts in the end.

\paragraph{Knowledge Graph}\label{kg}
We use \textit{ConceptNet} \citep{speer2017conceptnet}, a general-domain knowledge graph, as our structured knowledge source $G$. It has 799,273 nodes and 2,487,810 edges in total. Node embeddings are initialized using the entity embeddings prepared by \citet{feng-etal-2020-scalable}, which consists of four steps: (1) it first converts knowledge triples in the KG into sentences using pre-defined templates for each relation; (2) it then feeds these sentences into PLM to compute embeddings for each sentence; (3) after that, it extracts all token representations of the entity's mention spans in these sentences; (4) it finally mean pools over these representations and projects this pooled representation.

\paragraph{Implementation Details}
Our model is implemented using Pytorch and based on the Transformers Library \citep{wolf-etal-2020-transformers}. We fine-tune DeBERTa-V3-Large as the backbone pre-trained language model for \textsc{Decker}, and the hyper-parameter setting generally follows DeBERTa \citep{he2021debertav3}. 
We set the layer number of the R-GCN as 3, with a dropout rate of 0.1 applied to each layer. The number of retrieved facts is set to 5 due to the trade-off for computation resources. The maximum input sequence length is 256. The initial learning rate is selected in \{5e-6, 8e-6, 9e-6, 1e-5\} with a warm-up rate of 0.1. The batch size is selected in \{8, 16\}. We run up to 20 epochs and select the model that achieves the best result on the development dataset.

\subsection{Main Results}
Table \ref{results} presents the detailed results on two commonsense fact verification benchmarks: CREAK and CSQA 2.0. We compare our model with several baseline methods, which represent distinct knowledge-enhanced methods. UNICORN \citep{lourie2021unicorn} is instilled with external commonsense knowledge during the pre-training stage. GreaseLM \citep{zhang2022greaselm} integrates structured knowledge into models during the fine-tuning stage. RACo \citep{yu2022retrieval} incorporates unstructured knowledge by constructing a commonsense corpus on which its retriever is trained \footnote{RACo consists of two BERT-base models and T5-3B. The magnitude of the total parameter number depends largely on the latter, hence the sign of $\geq$ (greater than equal) is employed in Table \ref{results}.}.  Besides, we also compare our model with strong PLMs such as T5-3B \citep{t5}. 

The results indicate that our model \textsc{Decker} outperforms the strong baseline methods and achieves comparable results on the test set of CREAK. Besides, our model surpasses the current state-of-the-art model RACo on the contrast set of CREAK. Moreover, we observe that our model is lightweight and competitive without a considerable number of parameters and mixed data from multiple tasks during training, thus showing the strength and superiority of our model in various dimensions.

\begin{figure*}[!htb]
    \centering
    \includegraphics[width=1.0\textwidth]{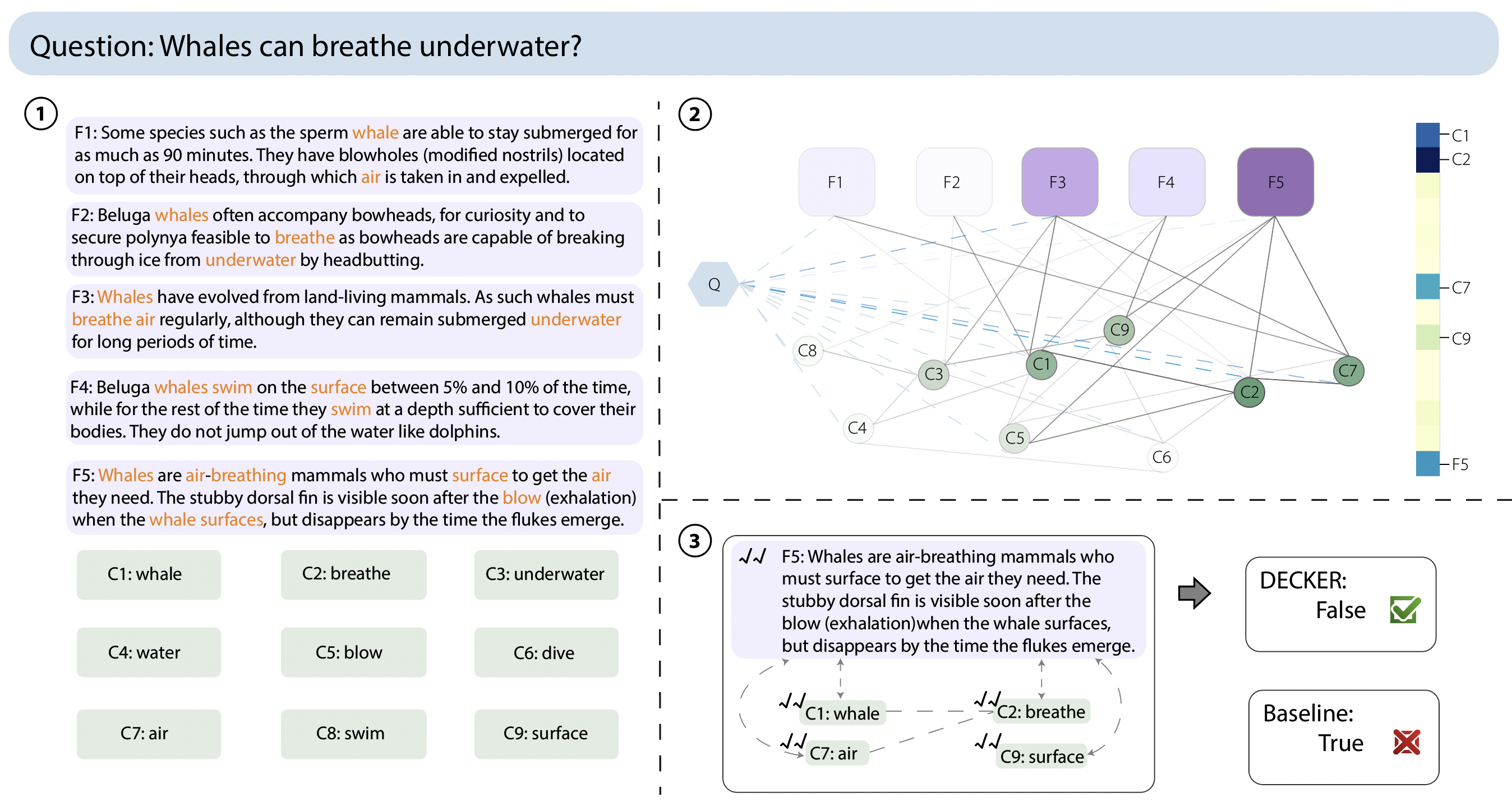}
    \caption{An example showing how our model works to achieve the correct answer, in which our baseline fails. Texts in purple denote facts and texts in green denote concepts.}
    \label{case}
\end{figure*}

\begin{table}[t]
	\centering
        \small
	\setlength{\tabcolsep}{8pt}
	{\begin{tabular}{ll}
		\toprule
		Model & Accuracy \\ 
		\midrule
            \textsc{Decker}  & \textbf{89.5}  \\
            \cdashline{1-2}[1pt/2pt]
            \multicolumn{2}{l}{\textbf{Knowledge Retrieval}}\\
        
            \quad w/o facts & 87.8($\downarrow 1.7$)   \\
            \quad w/o knowledge graph & 87.9($\downarrow 1.6$) \\
            \quad w/o both & 86.1($\downarrow 3.4$) \\
        
            \multicolumn{2}{l}{\textbf{Graph Construction}}\\
	    \quad w/o question node  & 89.3($\downarrow 0.2$)\\
            \quad w/o edge type & 87.6($\downarrow 1.9$) \\
            \quad w/o concept-to-fact edges & 88.1($\downarrow 1.4$)\\ 
            \quad w/o question-to-fact edges & 88.8($\downarrow 0.7$)\\ 
            \quad w/o concept-to-concept edges & 88.3($\downarrow 1.2$)\\ 
            \quad w/o question-to-concept edges & 89.1($\downarrow 0.4$)\\         
		\bottomrule
	\end{tabular}
	}
	\caption{Ablation study of our model for components in Knowledge Retrieval and Graph Construction modules on the CREAK development set.}
	\label{ablation}
\end{table}

\section{Analysis}
\subsection{Ablation Study}
We conduct a series of ablation studies under the same set of hyperparameters to determine the contributions of key components in our model. Results in Table \ref{ablation} demonstrate that the combination of heterogeneous knowledge and the components in our \textsc{Decker} are both non-trivial. Results in Table \ref{baseline} indicate that our \textsc{Decker} outperforms the baseline by a large margin. 
\paragraph{Knowledge Retrieval}
To investigate the effectiveness of knowledge combination, we discard the knowledge graph, facts and both. The resulting performances drop to 87.8\%, 87.9\%, and 86.1\% respectively, which reveals the necessity of fusing knowledge with different granularity.

\begin{table}[t]
	\centering
        \small
	\setlength{\tabcolsep}{10pt}
	{\begin{tabular}{lll}
		\toprule
		Model & CSQA2.0 & CREAK\\ 
            \midrule
            \text{DeBERTa}$_{large}$ & 67.9 & 86.1 \\
            \text{\textsc{Decker}} & 70.2($\uparrow 2.3$) & 89.5($\uparrow 3.4$) \\
		\bottomrule
	\end{tabular}
	}
	\caption{Results on the CSQA2.0 and CREAK development sets. The evaluation metric is accuracy (acc).}
	\label{baseline}
\end{table}

\begin{table}[t]
	\centering
        \small
	\setlength{\tabcolsep}{6pt}
	{\begin{tabular}{lcc}
		\toprule
		Model & Interaction &Accuracy\\ 
            \midrule
            \text{DeBERTa}$_\text{LARGE}$  & \okmark & 86.1  \\
            \quad w/ max pooling  & \ngmark & 87.5 \\
            \quad w/ mean pooling & \ngmark &86.7 \\
            \quad w/ attention pooling & \okmark &88.9 \\
            \quad w/ MHA pooling & \okmark &\textbf{89.5} \\
		\bottomrule
	\end{tabular}
	}
	\caption{Results of different pooling methods on the CREAK development set, MHA pooling denotes multi-head attention pooling for short.}
	\label{pooling}
\end{table}

\paragraph{Graph Construction}
One of the crucial components of our model is graph construction, where the integral graph contains three types of nodes and four types of edges. We ablate the question node and remove all the edges connected with it. The results show that the removal hurts the performance. Furthermore, we dive into the edge analysis. We first treat all edges as the same type instead of four types, which witnesses a significant drop in performance. Our intuition is that effective reasoning among heterogenous knowledge should attend to edge types because they symbolize the distinct emphases during reasoning. We then erase each kind of edge respectively. Notably, the absence of concept-to-fact edges degrades the performance badly, suggesting the necessity of double-checking between heterogeneous knowledge.

\subsection{Methods of Pooling}

During the period of aggregating the graph output, we analyze the influence of different pooling methods, including max pooling, mean pooling, attention pooling and multi-head attention pooling. These pooling methods can be divided into two categories: those  involving and those ignoring the interaction with the question. We compare the models with the same hyper-parameters on the development set of CREAK. Results in Table \ref{pooling} demonstrate that the interaction process promotes the model performance, which may reveal that the graph reasoning executes more on the information flow between different levels of knowledge and the augmented inquiry about the initial question implements a final refinement of enriched knowledge. As shown in Table \ref{pooling}, employing multi-head attention pooling presents the best performance.

\subsection{Interpretability: Case Study}
In order to further explore the mechanism and get more intuitive explanations of our model, we select a case from CREAK in which the baseline model fails but our model succeeds. In addition, we analyze the node attention weights related to the question induced in MHA mechanism. Figure \ref{case} shows that our \textsc{Decker} can well bridge the reasoning between heterogeneous knowledge, thus leading to better filtering the noisy material and maintaining the beneficial information. Concretely, given the claim \textit{whales can breathe underwater}, our model first extracts relevant structured and unstructured knowledge and then conducts reasoning over them. After reasoning, our model pays close attention to the concepts including \textit{breathe}, \textit{whale}, \textit{air}, \textit{surface} and the fact \textit{whales are air-breathing mammals who must surface to get the air they need}, as shown in the attention heatmap. We can see that our model has the capability of manipulating heterogeneous knowledge to answer the questions.


\section{Conclusion}
In this work, we propose \textsc{Decker}, a commonsense fact verification model that bridges heterogeneous knowledge and performs a double check based on the interactions between structured and unstructured knowledge. Our model not only uncovers latent relationships between heterogeneous knowledge but also conducts effective and fine-grained knowledge filtering of the knowledge. Experiments on two commonsense fact verification benchmarks (CSQA2.0 and CREAK) demonstrate the effectiveness of our approach. While most existing works focus on fusing one specific type of knowledge, we open up a novel perspective to bridge the gap between heterogeneous knowledge to gain more comprehensive and enriched knowledge in an intuitive and explicit way.

\section*{Limitations}

There are three limitations. First, our model requires the retrieval of relevant structured and unstructured knowledge from different knowledge sources, which can be time-consuming. Using cosine similarity over question and fact embeddings can be a bottleneck for the model performance. Second, our model focuses on rich background knowledge but might ignore some inferential knowledge, which can be acquired from other sources such as Atomic. Third, our model might not be applicable to low resources languages where knowledge graphs are not available.


\bibliography{anthology,custom}
\bibliographystyle{acl_natbib}

\appendix



\end{document}